\documentclass[journal,transmag]{IEEEtran}
\usepackage{graphicx}
\usepackage{amssymb}
\usepackage{subfigure}
\usepackage{makecell}
\usepackage{color}
\usepackage{multirow}
\usepackage{rotating}

\ifCLASSINFOpdf

\else

\fi

\usepackage{amsmath}
\usepackage{mathtools,nccmath}
\usepackage[linesnumbered,boxed,ruled,commentsnumbered]{algorithm2e}

\SetCommentSty{mycommfont}

\begin{document}

\title{Learning Representation over Dynamic Graph using Aggregation-Diffusion Mechanism}

\author{\IEEEauthorblockN{Mingyi Liu\IEEEauthorrefmark{1},
Zhiying Tu\IEEEauthorrefmark{1},
Xiaofei Xu\IEEEauthorrefmark{1}, 
Zhongjie Wang\IEEEauthorrefmark{1}}
\IEEEauthorblockA{\IEEEauthorrefmark{1}Faculty of Computing, Harbin Institute of Technology, Harbin, China \\
\\
\{liumy, tzy\_hit, xiaofei, rainy\}@hit.edu.cn}

\thanks{Manuscript received XXXXXX; XXXXXXXXX. 
Corresponding author: Zhongjie Wang (email: rainy@hit.edu.cn).}}

\markboth{Journal of \LaTeX\ Class Files,~Vol.~14, No.~8, August~2015}%
{Shell \MakeLowercase{\textit{et al.}}: Bare Demo of IEEEtran.cls for IEEE Transactions on Magnetics Journals}

\IEEEtitleabstractindextext{%
\begin{abstract}
Representation learning on graphs that evolve has recently received significant attention due to its wide application scenarios, such as bioinformatics, knowledge graphs, and social networks. The propagation of information in graphs is important in learning dynamic graph representations, and most of the existing methods achieve this by aggregation. However, relying only on aggregation to propagate information in dynamic graphs can result in delays in information propagation and thus affect the performance of the method. To alleviate this problem, we propose an aggregation-diffusion (AD) mechanism that actively propagates information to its neighbor by diffusion after the node updates its embedding through the aggregation mechanism. In experiments on two real-world datasets in the dynamic link prediction task, the AD mechanism outperforms the baseline models that only use aggregation to propagate information. We further conduct extensive experiments to discuss the influence of different factors in the AD mechanism.

\end{abstract}

}

\maketitle

\IEEEdisplaynontitleabstractindextext

\IEEEpeerreviewmaketitle

\section{Introduction}\label{sec:intro}
Representation learning on graph structured data has recently received significant attention due to its wide application scenarios in various domains such as social networks, knowledge graphs, and bioinformatics. Recently, graph neural networks (GNNs)\cite{scarselli2008graph, kipf2016semi,wang2016structural} have been applied to learn high-dimensional and non-Euclidean graph information efficiently. However, most of the existing GNNs are designed for static graphs. Graphs tend to evolve in real-world application scenarios. For example, a new friendship will be established between people in social networks, and human interaction in social networks changes during an epidemic.

Dynamic graphs can be represented in discrete or continuous form\cite{skarding2020foundations}, and in this paper, we focus on the continuous form representation, where no temporal aggregation is applied on the graph. Temporal point process based models\cite{trivedi2019dyrep,knyazev2019learning,han2020graph,trivedi2017know} are emerging to address representation learning for dynamic networks with continuous representation. The temporal point process (TPP) is modeled by events $o^t = (u, v, t, k)$ where $u$ and $v$ are the interacting nodes, $t$ is the time of the event, and $k$ is the category of this event.
\begin{figure}[htbp]
	\centering
	\includegraphics[width=\linewidth]{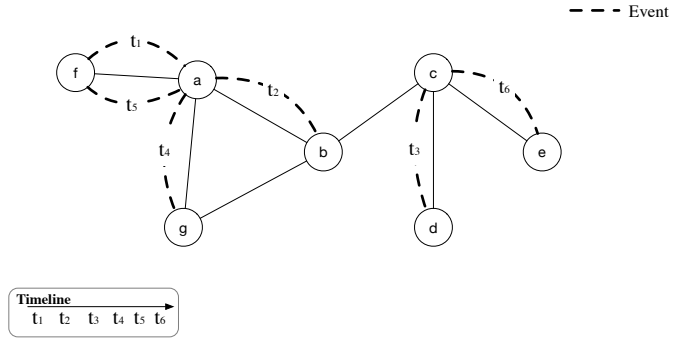}
    \caption{An example of problems with information propagation based on aggregation mechanisms.} \label{fig:delay_issuse}
\end{figure}

The TPP is parameterized by an RNN. When an event occurs, the previous embeddings of neighbors of the interacting node are aggregated and fed into this RNN to update the interacting node. This node update mechanism assumes that long-distance structural information and timely temporal information can be propagated to interacting nodes by aggregation mechanisms. However, this assumption does not always hold. Fig.~\ref{fig:delay_issuse} shows an example of problems with information propagation based on aggregation mechanisms: 1) Suppose the current time is $t_6$, and an event occurs between node $c$ and $e$. Then $c$ can get the information of node $a$ at time $t_1^+$ by aggregating the information of its neighbor $b$, since the information of $a$ at time $t_1^+$ is propagated to $b$ through aggregation when the event occurs at time $t_2$. By serving node $c$ as a bridge, node $e$ can also access the information of node $a$ at time $t_1^+$. However, before $t_6$, the information of $a$ is updated twice at $t_4$ and $t_5$, so $c$ and $e$ receive very lagged information of $a$. 2) $b$ is hub-node for $\{a, f, g\}$ and $\{c, d, e\}$, which means that no information can be exchanged between $\{a, f, g\}$ and $\{c, d, e\}$ unless some event happens on $b$, or some event happens between $\{a, f, g\}$ and $\{c, d, e\}$.

To alleviate this limitation, we introduce the aggregation-diffusion (AD) mechanism into TPP-based dynamic network representation learning. The core operation of the AD mechanism is that the interacting nodes actively diffuse the changes to their neighbors after their information is updated by aggregating information about their neighbors. 
The main intuition behind this mechanism is twofold: 1) \textbf{aggregation}: a node's information is affected by its neighbors; 2) \textbf{diffusion:} changes in the node itself should be sensed by its neighbors in time, even if there is no explicit event occurring. We use two representative TPP-based models, DyRep\cite{trivedi2019dyrep} and LDG\cite{knyazev2019learning}, as backbone model to demonstrate the effectiveness and scalability of the AD mechanism. On two dynamic graph datasets, Social Evolution\cite{madan2011sensing} and Github\footnote{https://www.gharchive.org/}, the model with AD mechanism produce significant performance improvement on both Mean Average Rank (MAR) and HITS@10 compared to the original DyRep(LDG) model. We further conduct extensive experiments to discuss the influence of different factors in the aggregation-diffusion mechanism.

The remainder of this paper is organized as follows: In Section \ref{sec:related_work}, we introduce related work. In Section \ref{sec:backgroud}, we describe relevant details of the DyRep and LDG. In Section~\ref{sec:ad}, we explain AD mechanism in detail. In Section~\ref{sec:settings}, we give the details of the experiment settings. In Section~\ref{sec:results}, we present the experiment results and discuss different impact factors of the AD mechanism. In the final section, we present the conclusion.

\section{Related Word}\label{sec:related_work}

\subsection{Discrete Dynamic Embedding Approaches}
Discrete dynamic embedding approaches treat dynamic graph as a sequence of graph snapshots. Most of discrete dynamic embedding approaches\cite{seo2018structured,narayan2018learning,niepert2016learning,chen2019lstm,zheng2020mathnet} focus on the learning representations of entire dynamic graphs rather than node representations. Some approaches\cite{sanchez2018graph,chang2016compositional,manessi2020dynamic,jin2019recurrent} are starting to focus on the dynamic representation at node level, they encode each graph snapshot using static embedding approaches\cite{liben2007link,hisano2018semi,hamilton2017inductive,tang2015line,grover2016node2vec,kipf2016semi} to embed each node, and then combines some time-series models (e.g. LSTM\cite{hochreiter1997long}, RNN\cite{sherstinsky2020fundamentals}) for per node to model the discrete dynamic.

\subsection{Continuous Dynamic Embedding Approaches}
Currently, continuous dynamic embedding approaches are divided into two main categories: RNN based approaches and temporal point processes based approaches\cite{skarding2020foundations}. 

In RNN based approaches, the embedding of interacting nodes is updated by the a RNN based architecture according to the historical information of itself. Representative works of this type of approach are JODIE\cite{kumar2019predicting}, TGN\cite{rossi2020temporal} and Streaming graph neural network\cite{ma2020streaming}. JODIE\cite{kumar2019predicting} are designed for user-item interaction networks, which uses two RNN to maintain the embedding of each node. With on RNN for users and another one for items. Instead of keeping the embedding of node directly, TGN\cite{rossi2020temporal} calculates the embedding of node at different time by introducing message and memory mechanisms. The architecture of Streaming graph neural network\cite{ma2020streaming} is consist of two components: 1) update component; and 2) propagation component. The update component is used to update the embedding of nodes involved in an event and the propagation component propagates the event to the involved nodes neighbors. The process of our proposed mechanism is similar to \cite{ma2020streaming}, but there are several significant differences: 
\begin{itemize}
    \item The aggregated information is different, \cite{ma2020streaming} aggregate the node's own historical information, while we focus more on aggregating neighbors information, which is more effective in TPP based models\cite{trivedi2019dyrep,knyazev2019learning,han2020graph}.
    \item The propagation information, propagation objects, and propagation methods are different.
    \item \cite{ma2020streaming} focus more on the architecture of the neural network, while we focus on a mechanism that can be adapted to existing TPP based approaches.
    \item We discuss the influence of different factors on the aggregated-diffusion mechanism in more detail.
\end{itemize}

Know-Evolve\cite{trivedi2017know} is the pioneer in bringing the temporal point processes\cite{cox2020multivariate} to dynamic graph representation learning, which models temporal knowledge graph as multi-relational timestamped edges by parameterizing a TPP by a deep recurrent architecture. DyRep\cite{trivedi2019dyrep} is the successor of Know-Evolve. DyRep extends Know-Evolve by using TPP to model long-term events (topological evolution) and short-term events (node communication) and introducing aggregation mechanisms. LDG\cite{knyazev2019learning} argues long-term events are often specific by humans, and can be suboptimal and expensive to obtain. LDG use Neural Relational Inference (NRI) model\cite{kipf2018neural} to infer the type of events on the graph and replaces the self-attention originally used in DyRep by generating a temporal attention matrix to better aggregate neighbor information. GHN\cite{han2020graph} is another TPP based approach, which uses an adapted continuous-time LSTM for Hawkes process\cite{mei2017neural}. Similar to Know-Evolve, GHN is specifically designed for knowledge graphs.

In this paper, we choose DyRep and LDG as our backbone model for the following reasons:
\begin{itemize}
    \item DyRep and LDG are universal and  not specifically designed for knowledge graphs or user-item graphs.
    \item DyRep and LDG can model realistic long-term events and short-term events, which can not be provided by other models.
    \item DyRep and LDG have an obvious aggregation mechanism when interacting nodes are updating.
\end{itemize}
It is important to note that theoretically any model that updates node embeddings in continuous dynamic graphs using the aggregation mechanism can be extended using the aggregation-diffusion mechanism.

\section{Background: DyRep \& LDG}\label{sec:backgroud}
In this section, we describe relevant details of the DyRep and LDG model. \textbf{We strongly recommend readers to read the original article\cite{trivedi2019dyrep,knyazev2019learning} for more details to better understand the details of how DyRep and LDG models work.} LDG and DyRep represent the evolution of dynamic graph as two distinct processes:
\begin{itemize}
    \item \textbf{Long-term association ($k = 1$).} This is also called ``dynamic of graph'', in which new nodes or edges are added resulting in a change in the topology of the graph.
    \item \textbf{Short-term communication ($k = 0$)}. This is also called ``dynamic on graph'', in which interaction between nodes leads to temporary information flow between these nodes\cite{farine2017dynamics,artime2017dynamics}. And this process does not change the topology of the graph.
\end{itemize}

\subsection{Node update}
\begin{figure*}
    \centering
    \includegraphics[width=\linewidth]{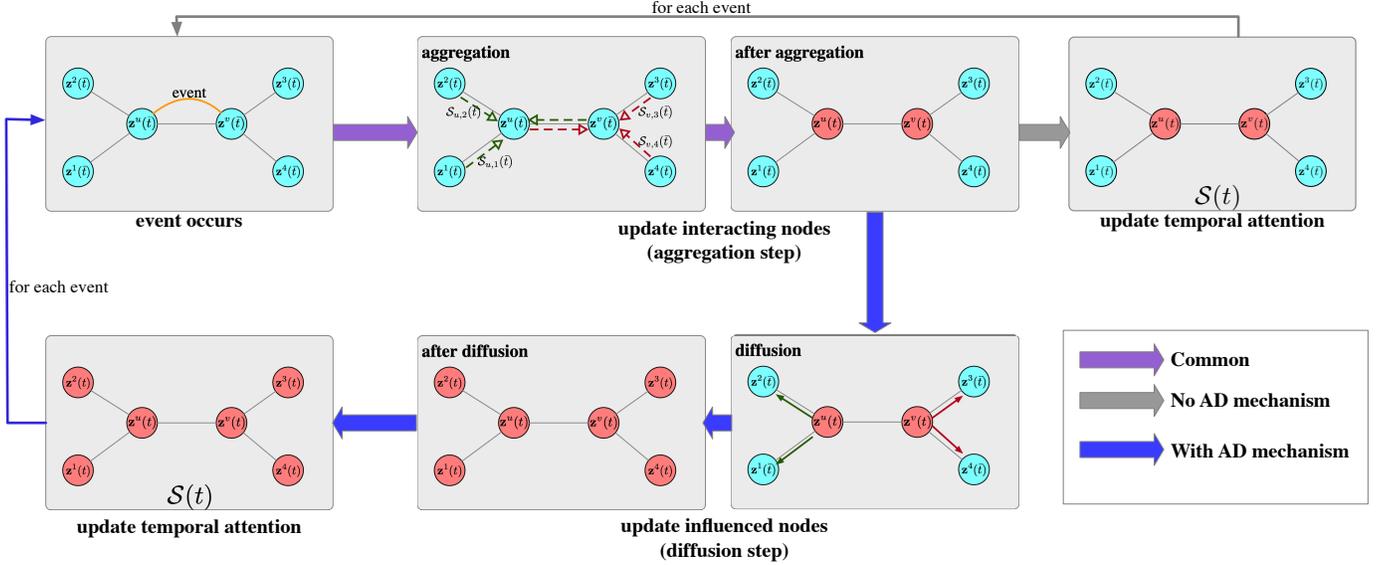}
    \caption{Overview of node embedding update process with AD mechanism and without AD mechanism.}
    \label{fig:node_update}
\end{figure*}
The node update mechanism is same for both DyRep and LDG. When an event $o=(u, v, t, k)$ occurs between interacting nodes $u$ and $v$ will cause their node embeddings $\mathbf{z}^{u}, \mathbf{z}^{v} \in \mathbb{R}^d$ to be updated and subsequently update the temporal attention $\mathcal{S} \in \mathbb{R}^{N \times N}$. The update process are show in Fig.~\ref{fig:node_update}.

In particular, when an event occurs, the embedding of participating node $u$ is updated based on the three terms of \textbf{Self-propagation}, \textbf{Exogenous Drive} and \textbf{Attention-based Aggregation}. Specially, for an event of node $u$ at time $t$, updating $\mathbf{z}^{u}$ as:
\begin{equation}\label{eq:interacting_update}
    \mathbf{z}^{u}(t) = \sigma(\underbrace{\mathbf{W}^{s}\mathbf{h}^{u}_{s}(\bar{t}))}_{\textbf{Aggregation}} + \underbrace{\mathbf{W}^{r}\mathbf{z}^{u}(\bar{t^u})}_{Self-propagation} + \underbrace{\mathbf{W}^{t}(t-\bar{t^u})}_{\text{Exogenous Drive}})
\end{equation}
where $\mathbf{W}^{s} \in \mathbb{R}^{d \times d}$, $\mathbf{W}^{r} \in \mathbb{R}^{d \times d}$ and $\mathbf{W}^{t} \in \mathbb{R}^d$ are learned parameters used to control the effect of above-mentioned three terms on the computation of node embedding, respectively. $\sigma(\cdot))$ is a nonlinear function. $\mathbf{z}^{u}(\bar{t^u})$ is the previous representation of node $u$. $\bar{t}$ denotes the time point just before current event time $t$ and $\bar{t^u})$ represent the time point of last event involved $u$. $\mathbf{h}^{u}_{s}(\bar{t}) \in \mathbb{R}^d$ is the output representation obtained from the aggregation of node $u$'s neighbors $\mathcal{N}_u^a$:
\begin{equation}
    \mathbf{h}^{u}_{s}(\bar{t}) =  Agg(\text{softmax}(\mathcal{S}_u(\bar{t}))_r(\mathbf{W}^{h}\mathbf{z}^{r}(\bar{t^u})),  \forall r \in \mathcal{N}_u^{a}))
\end{equation}
where $Agg(\cdot)$ is an aggregation function and $\mathbf{W}^{h} \in \mathbb{R}^{d \times d}$ are learned parameters. The amount of information propagated from node $u$'s neighbors is controlled by temporal attention $\mathcal{S}_u(\bar{t}))$, which is updated by a hard-coded algorithm in DyRep and learned in LDG. It should be noted the update of $\mathcal{S}$ is affected by nodes' embedding. In DyRep, temporal attention $\mathcal{S}(t)$ relies on adjacency matrix $\mathbf{A}(\bar{t})$ and temporal attention $\mathcal{S}(\bar{t})$ at previous time step and conditional intensity $\lambda_{k}^{u,v}(t)$ of event record $o = (u, v, t, k)$:
\begin{equation}
    \mathcal{S}(t) = f_S(\mathbf{A}(\bar{t}), \mathcal{S}(\bar{t}), \lambda_{k}^{u,v}(t))
\end{equation}
where $f_S$ is the attention update function in DyRep\cite{trivedi2019dyrep}. Conditional intensity $\lambda_{k}^{u,v}(t)$ models the occurrence of event $o=(u, v, k, t)$ between $u$ and $v$ at time $t$:
\begin{equation}
    \lambda_{k}^{u,v}(t) = \psi_k\log \left(1 + \exp \left\lbrace \frac{\mathbf{\omega}_k^T[\mathbf{z}^u(\bar{t});\mathbf{z}^v(\bar{t})]}{\psi_k}\right \rbrace \right)
\end{equation}
where $\psi_k$ is trainable scalar parameter, which denotes the rate of events arising from a corresponding process, and $\mathbf{\omega}_k \in \mathbb{R}^{2d}$ is designed to learn time-scale specific compatibility. $[;]$ denotes concatenation.

In LDG\cite{knyazev2019learning}, they replace the hard-coded node update algorithm $f_S$ with a learnable bilinear encoder $f_S^{enc}$, which is a two pass progress to ensure temporal attention $\mathcal{S}(t)$ depends on node embeddings at previous time step:
\begin{equation}
    \mathcal{S}(t) = f_S^{enc}(\mathbf{z}(t-1))
\end{equation}

\section{Aggregation-Diffusion Mechanism}\label{sec:ad}
\begin{algorithm}[h]
\caption{Update Node Embedding with Aggregation-Diffusion Mechanism}\label{ag:update_node}
\SetKwInOut{Input}{Input}
\SetKwInOut{Output}{Output}

\Input{
    Event record $o = (u, v, t, k)$ \; \\
    All node embeddings of previous time $\mathbf{z}(\bar{t})$ \; \\
    Most recently updated $\mathbf{A}(\bar{t})$ and $\mathcal{S}(\bar{t})$ \; \\
    Trainable parameters $\mathbf{W}^{s}$, $\mathbf{W}^{r}$, $\mathbf{W}^{t}$, $\mathbf{W}^{h}$ and $\mathbf{W}^{d}$\; \\}

\Output{
    Updated node embeddings $\mathbf{z}(t)$ \;
}
\tcc{Aggregation step}
\For{\textbf{each} $j \in \{u,v\}$}{
    $\mathcal{N}_j^{a} \gets \{r: \mathbf{A}_{r,j}(\bar{t}) > 0 \}$ \\
    \tcc{Aggregate information from all one-hop neighbors. Discussed in \textbf{Section~\ref{sec:who}}}
    $\mathbf{h}^{j}_{s}(\bar{t}) \gets \text{\quad \quad} Agg(\text{softmax}(\mathcal{S}_j(\bar{t}))_r(\mathbf{W}^{h}\mathbf{z}^{r}(\bar{t^j})),  \forall r \in \mathcal{N}_j^{a}))$ \\
    \tcc{Update interacting node's embedding}
    $\mathbf{z}^{j}(t) \gets \sigma(\mathbf{W}^{s}\mathbf{h}^{j}_{s}(\bar{t})) + \mathbf{W}^{r}\mathbf{z}^{j}(\bar{t}) + \mathbf{W}^{t}(t-\bar{t^j}))$
}

\tcc{Diffusion step}
\For{\textbf{each} $j \in \{u,v\}$}{
    \tcc{Generating diffusion message. Discussed in \textbf{Section~\ref{sec:message}}}
    $\mathbf{m}^{j}(t) \gets Generator(\mathbf{z}(t), \mathbf{z}(\bar{t}), o)$ \\
    \tcc{Selecting candidate diffusion nodes. Discussed in \textbf{Section~\ref{sec:who}} and \textbf{Section~\ref{sec:hop}}}
    $\mathcal{N}_j^{d} \gets SelectCandidate(\mathbf{A}(\bar{t}))$ 
    \tcp{$\mathcal{S}(\bar{t})$ for LDG} 
    \For{\textbf{each} $r \in \mathcal{N}_j^{d}$}{
        \tcc{Update diffused node's embedding. Discussed in \textbf{Section~\ref{sec:attn}}}
        $\mathbf{z}^{r}(t) \gets \sigma(\mathbf{z}^{r}(\bar{t}) + q_{j,r}(\bar{t})\mathbf{W}^{d}\mathbf{m}^{j}(t))$
    }
}
\Return $\mathbf{z}(t)$
\end{algorithm}
In this paper, we extend DyRep and LDG by using a node embedding update algorithm with aggregation-diffusion (AD) mechanism. As mentioned in Section~\ref{sec:intro}, the main intuition behind AD mechanism is straightforward: first, the node's information is affected by its neighbors, which is aggregation part; second, changes in the node itself should be propagated to its neighbors proactively and in a timely manner, which is diffusion part.

Algorithm~\ref{ag:update_node} gives the pseudo-code for node embedding update with AD mechanism. The algorithm consists of two steps: 1) an aggregation step, which is used to update the embeddings of the nodes directly involved in a event and has been explained in Section~\ref{sec:backgroud}; 2) a diffusion step, which is used to update the embedding of other nodes that may be affected and will be discussed in this section.

The diffusion step mainly consists of diffusion message generation, diffusion node selection and update of diffused nodes.

\textbf{Diffusion Message Generation.} It determines what kind of message the interacting node of an event will propagate to its neighbors. We model the message in three ways. The most straightforward way is interacting node $u$ diffuses its updated embedding $\mathbf{z}^u(t)$ outward and the formulation is as follows:
\begin{equation} \label{eq:message_node}
    \mathbf{m}^{u}(t) = \mathbf{z}^u(t)
\end{equation}
An interacting node can also diffuse outward changes in itself rather than just current state:
\begin{equation}\label{eq:message_delta}
    \text{delta:} \quad \mathbf{m}^{u}(t) = \mathbf{z}^u(t) - \mathbf{z}^u(\bar{t})
\end{equation}
There is also a way to diffuse the impact of the event outward:
\begin{equation}\label{eq:message_edge}
    \text{edge:} \quad \mathbf{m}^{u}(t) = \sigma(\mathbf{W}^{1}\mathbf{z}^u(t) + \mathbf{W}^{2}\mathbf{z}^v(t))
\end{equation}
where $v$ is another interacting node in the event, $\mathbf{W}^{1}, \mathbf{W}^{2} \in \mathbb{R}^{d \times d}$ are trainable parameters.

These three different approaches to diffusion message have their own advantages and disadvantages, which will be discussed in detail in Section~\ref{sec:message}.

\textbf{Diffusion Node Selection.} This is designed to select the nodes that will receive the diffusion information. 

In this paper, we choose the 1-hop neighbors of the interacting node as the diffusion nodes. We do not diffuse more hops because more hops will result in a significant decrease in training speed but not a significant performance improvement or even a decrease in performance due to the introduction of noise. We will discuss the impact of diffusion hops in detail in Section~\ref{sec:hop}.

Specially, when diffusing the message $\mathbf{m}^{u}(t)$ generated by node $u$, we will avoid involving another interacting node $v$ in the event, because $v$ has already obtained information about $u$ through aggregation step, and repeatedly obtaining information through diffusion will lead to a negative effect, which will be discussed in Section~\ref{sec:who}. Therefore, the formulation for diffusion node selection is as follows:
\begin{equation}\label{eq:who_neighbor}
    \mathcal{N}_u^{d} = \{r: \mathbf{A}_{r,u}(\bar{t}) > 0 \text{ and } r \ne v\}
\end{equation}
It should be noted, for LDG, we use $\mathcal{S}(\bar{t})$ to replace $\mathbf{A}(\bar{t})$, because $\mathbf{A}$ is not maintained in LDG.

In addition, we also tried to mask the aggregation/diffusion nodes randomly and temporally, which are also discussed in Section~\ref{sec:who}.

\textbf{Update of Diffusion nodes.} The diffusion nodes will update their embeddings based on their previous embedding and the diffusion message:
\begin{equation}\label{eq:d_u}
    \mathbf{z}^{r}(t) = \sigma(\mathbf{z}^{r}(\bar{t}) + q_{u,r}(\bar{t})\mathbf{W}^{d}\mathbf{m}^{u}(t)), \forall r \in \mathcal{N}_u^{d}
\end{equation}
where $\mathbf{W}^{d} \in \mathbb{R}^{d \times d}$ is a trainable parameters, and $q_{u,r}(\bar{t})$ is used to control the strength of diffusion from $u$ to $r$. In this paper we discuss two methods of calculating $q_{u,r}(\bar{t})$. One is \textit{uniform}, where all values are equal to $1$:
\begin{equation}\label{eq:uniform}
    \text{uniform:} \quad q_{u,r}(\bar{t}) = 1, \forall r \in \mathcal{N}_u^{d}
\end{equation}
Another one is \textit{attention}, which uses temporal attention $\mathcal{S}(\bar{t})$ to calculate $q_{u,r}(\bar{t})$:
\begin{equation}\label{eq:attention}
    \text{attn:} \quad q_{u,r}(\bar{t}) = \frac{\exp(\mathcal{S}_{u,r}(\bar{t}))}{\sum_{r'\in \mathcal{N}_u^{d}}\exp(\mathcal{S}_{u,r'}(\bar{t}))}, \forall r \in \mathcal{N}_u^{d}
\end{equation}
This will be discussed in Section~\ref{sec:attn}.

\section{Experiment Settings}\label{sec:settings}
\subsection{Datasets \& Metrics}
\begin{table}[htpp]
\centering
\caption{Dataset Statistics for Social Evolution and Github. }\label{tab:dataset}
\begin{tabular}{lcc}
\hline
                      & \begin{tabular}[c]{@{}c@{}}SOCIAL\\EVOLUTION\end{tabular} & GITHUB \\ \hline
\#Nodes                & 83               & 284    \\
\#Initial Associations & 575              & 149    \\
\#Final Associations   & 708              & 710    \\
\#Train Event          & 43,834           & 11,644 \\
\#Test Event           & 10,535           & 9,082  \\
\hline
\end{tabular}
\end{table}

We evaluate the AD mechanism on two real world dynamic graph datasets, Social Evolution\cite{madan2011sensing} and Github, which are also the dataset used in DyRep\cite{trivedi2019dyrep} and LDG\cite{knyazev2019learning}. The statistical results for Social Evolution and Github are presented in Table~\ref{tab:dataset}.

\begin{table*}[!htbp]
\centering
\caption{Performance comparison of whether to use AD machanism.  \textcolor{blue}{\textbf{Blue bolded}} results denote best performance for DyRep-based model, \textcolor[rgb]{0.502,0,0.502}{\textbf{Purple bolded}} results denote best performance for LDG-based model.}\label{tab:overview}

\begin{tabular}{ccccccccc} 
\hline
\multirow{2}{*}{MODEL} & \multicolumn{4}{c}{SOCIAL EVOLUTION}                                                                           & \multicolumn{4}{c}{GITHUB}                                                                                       \\ 
\cline{2-9}
                       & MAR                                           & HIT@10                                         & SPEED & Epoch & MAR                                            & HIT@10                                         & SPEED & Epoch  \\ 
\hline
DyRep                  & 13.88                                         & 0.486                                          & 1x    & 5     & 117.83                                         & 0.165                                          & 1x    & 5      \\
DyRep-self             & 20.61                                         & 0.141                                          & 0.9x  & 5     & 130.99                                         & 0.160                                          & 0.9x  & 5      \\
DyRep-D-base           & 6.74                                          & 0.897                                          & 4x    & 1     & 85.51                                          & \textbf{\textcolor{blue}{0.287}}               & 4x    & 2      \\
DyRep-AD-base          & \textcolor{blue}{\textbf{6.28}}               & \textcolor{blue}{\textbf{0.907}}               & 4x    & 1     & \textcolor{blue}{\textbf{81.25}}               & 0.262                                          & 4x    & 1      \\ \hline \hline
LDG                    & 13.06                                         & 0.448                                          & 40x   & 5     & 64.64                                          & 0.276                                          & 40x   & 2      \\
LDG-self               & 13.75                                          & 0.479                                           & 40x  & 5  & 59.43                                          & 0.290                                          & 40x   & 2      \\
LDG-D-base             & 6.94                                          & 0.902                                          & 40x   & 5     & 51.56                                          & 0.462                                          & 40x   & 3      \\
LDG-AD-base            & \textbf{\textcolor[rgb]{0.502,0,0.502}{6.40}} & \textbf{\textcolor[rgb]{0.502,0,0.502}{0.918}} & 40x   & 5     & \textbf{\textcolor[rgb]{0.502,0,0.502}{51.49}} & \textbf{\textcolor[rgb]{0.502,0,0.502}{0.480}} & 40x   & 2      \\
\hline
\end{tabular}
\end{table*}

\textbf{Social Evolution\cite{madan2011sensing}.} This dataset is released by MIT Human Dynamics Lab, which consists of over $2M$ events $o=(u, v, t, k)$. Follow \cite{trivedi2019dyrep}, we treat Proximity, Calls and SMS records between users as communication events (short-term events, $k=1$) and all Close Friendship records between users are treated as association events (long-term events, $k=0$). Follow \cite{knyazev2019learning}, Proximity records are filted by the probability that record occurred, because the number of Proximity records is too large and contains a lot of noise. The Social Evolution data is collected from Jan 2008 to June 2009. Similar to \cite{trivedi2019dyrep} and \cite{knyazev2019learning}, we use the association events between users from Jan 2008 to Sep 10, 2008 to initialize the graph, and events from Sep 11, 2008 to April 2009 is used as training set, and events after May 2009 is used as test set. After pre-processing, the dataset contains $83$ nodes, the training set contains $43K$ events, and the test set contains $10K$ events.

\textbf{Github.} This dataset is released by Github Archive. The original dataset contains over $12K$ nodes and $600K$ events and compared to Social Evolution is a large graph with sparse events. Since LDG requires sufficient interactions between nodes to train temporal attention $\mathcal{S}$, we extract a dense subgraph following the processing in \cite{knyazev2019learning}. Follow \cite{trivedi2019dyrep}, we treat Follow records between users as association events and other records are treated as communication records. We use the association events in 2011-2012 to initialize the graph, and events from Jan 1, 2013 to Sep 30, 2013 is used as training set, and events from Oct 1, 2013 to Dec 31, 2013 is used as test set.  After pre-processing, the dataset contains $284$ nodes, the training set contains $11K$ events, and the test set contains $9K$ events.

During the test, for a given event $(u, ?, t, k)$ or $(?, v, t, k)$, we compute the conditional density of known node with all other nodes and rank them. Same as \cite{trivedi2019dyrep} and \cite{knyazev2019learning}, we report Mean Average Ranking (MAR) and HIT@10.

\subsection{Implementation Details}
For LDG and DyRep we directly use the code\footnote{https://github.com/uoguelph-mlrg/LDG} provide by \cite{knyazev2019learning}. We only modify the node update function $update\_node\_embed$ in the code to add the AD mechanism into LDG and DyRep.

The hyper parameter setting of the experiments are also consistent with those of LDG. We use the Adam optimizer\cite{kingma2014adam} with the learning rate set to $0.0002$. The hidden units $d$ per layer is set to 32. Gradient clipping is used to avoid gradient explosion, and the clipping value is set to 100. We do not use dropout and batch size is set to 200. We train for 5 epochs. We run each experiment 10 times and report the average results.

\section{Results \& Discussion}\label{sec:results}
\subsection{Overview}

In this section we give an overview performance comparison between models using the AD mechanism and not using. The AD mechanism used in this section is \textit{*-AD-base}, which follows the principle of simplicity, specifically, using Eq.~\ref{eq:message_node} in generating diffusion message, considering only 1-hop neighbors, using Eq.~\ref{eq:who_neighbor} in selecting the diffusion nodes, and using a uniform strength $q$ (Eq.~\ref{eq:uniform}). We also conduct experiments on models that only use diffusion step (\textit{*-D-base}) as well as neither diffusion nor aggregation step (\textit{*-self}). Table~\ref{tab:overview} shows the performance comparison.

On Social Evolution, compared with the baseline \textbf{DyRep}, \textbf{DyRep-AD-base} reduced MAR from $13.88$ to $6.28$ and improved HIT@10 from $0.468$ to $0.907$; compared with baseline \textbf{LDG}, \textbf{LDG-AD-base} reduced MAR from $13.06$ to $6.40$ and improved HIT@10 from $0.448$ to $0.918$.

On Github, compared with the baseline \textbf{DyRep}, \textbf{DyRep-AD-base} reduced MAR from $117.83$ to $81.25$ and improved HIT@10 from $0.165$ to $0.262$; compared with baseline \textbf{LDG}, \textbf{LDG-AD-base} reduced MAR from $64.64$ to $51.49$ and improved HIT@10 from $0.276$ to $0.480$.

We find that the performance improvement from the diffusion step is significantly higher than that from the aggregation step, which may be due to the fact that both aggregation and diffusion are essentially for information delivery, and the number of nodes that acquire new information in the aggregation process (1 node) is less than that in the diffusion process (node's neighbors).

\begin{figure*}[!htbp]
    \centering
    \subfigure[MAR on Social Evolution]{
    \includegraphics[width=0.45\linewidth]{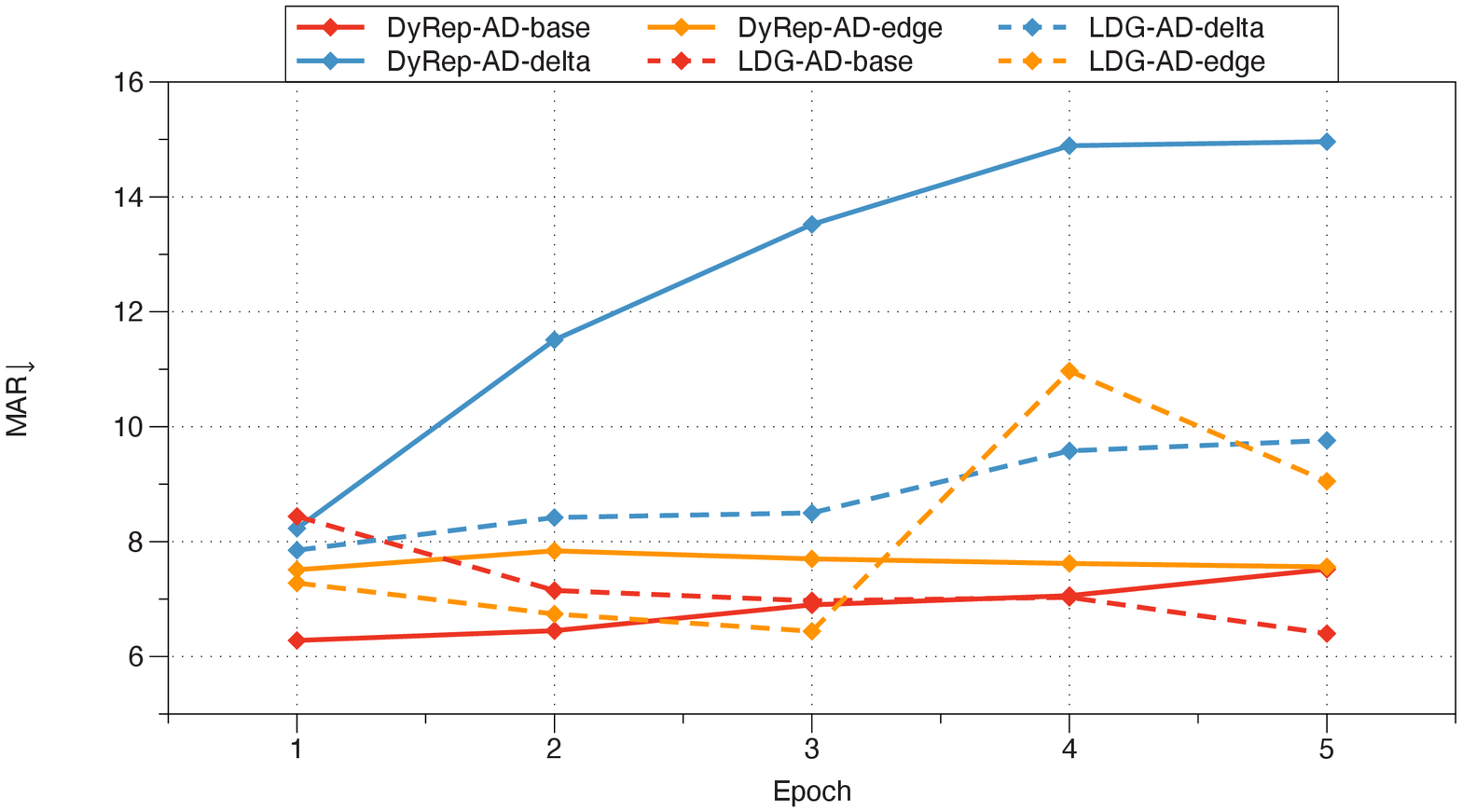}
    }
    \subfigure[MAR on Github]{
    \includegraphics[width=0.45\linewidth]{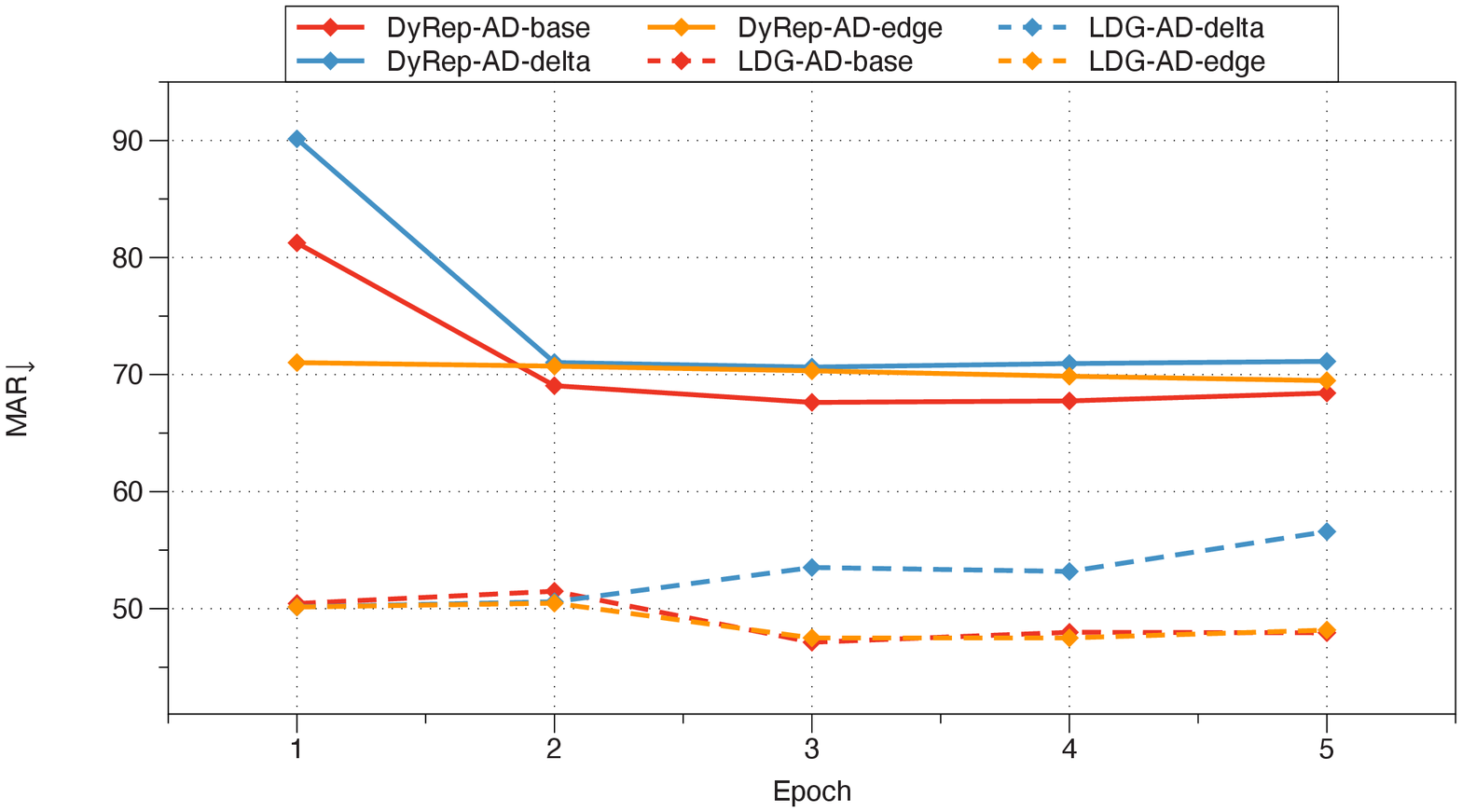}
    }
    \subfigure[HIT@10 on Social Evolution]{
    \includegraphics[width=0.45\linewidth]{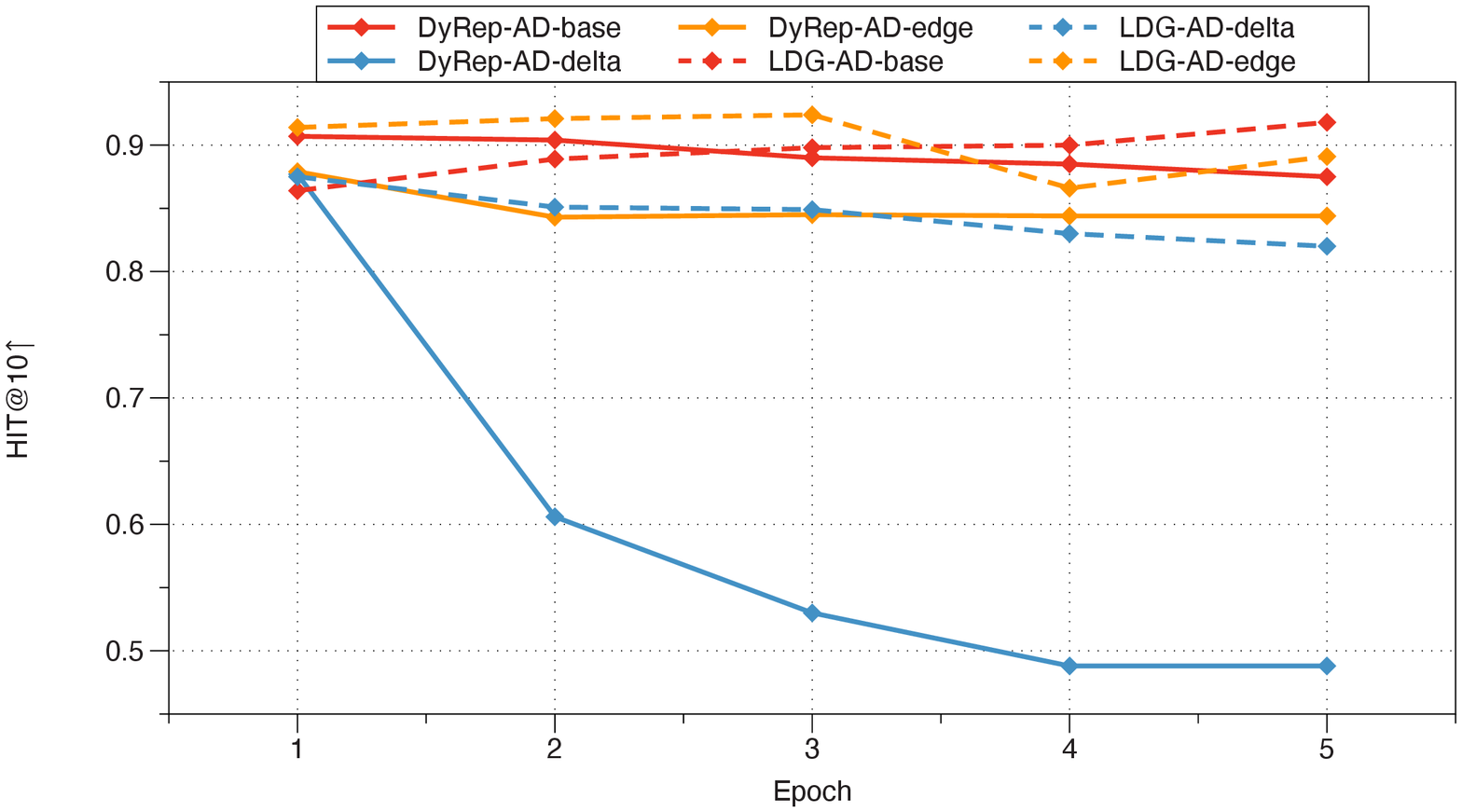}
    }
    \subfigure[HIT@10 on Github]{
    \includegraphics[width=0.45\linewidth]{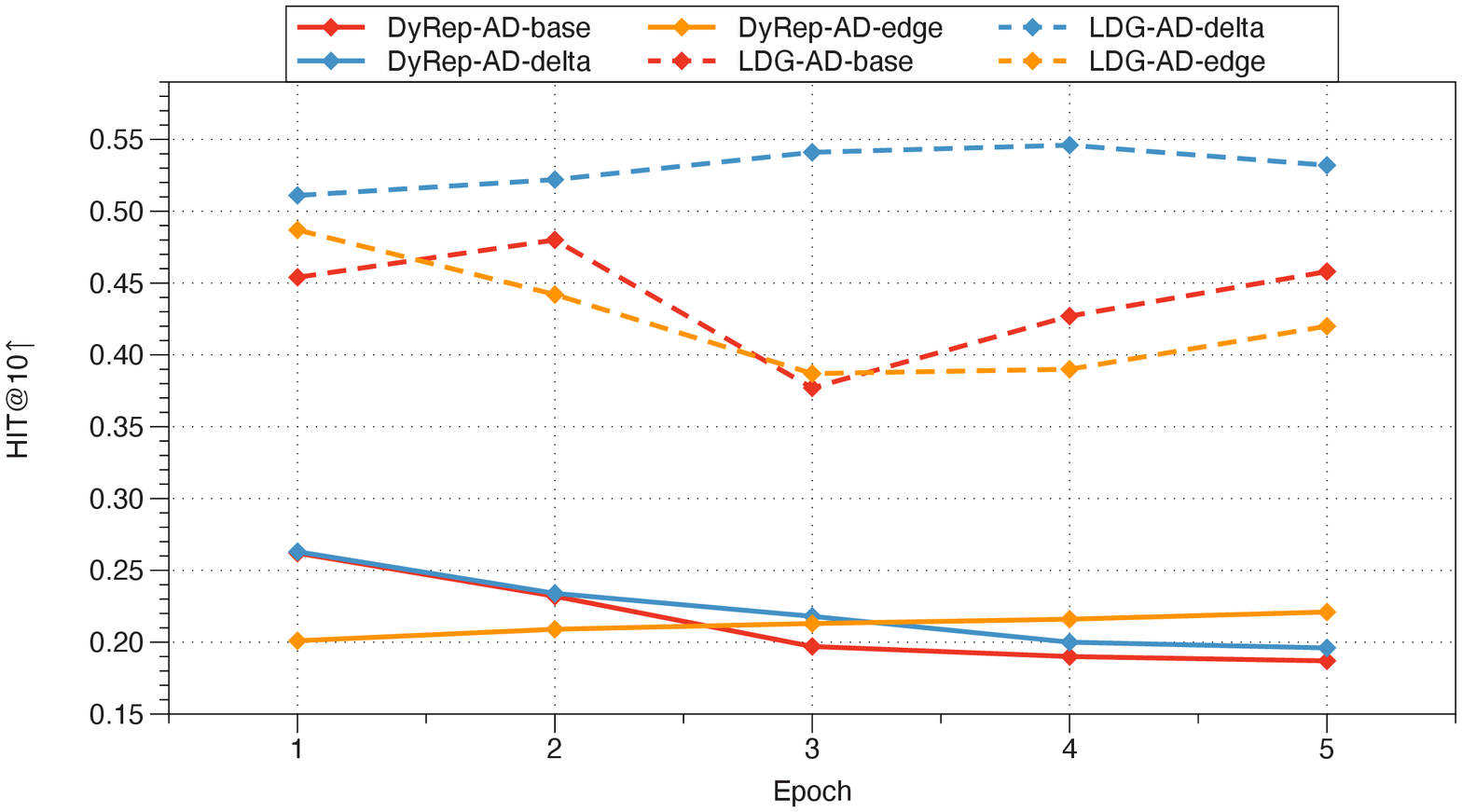}
    }
    \caption{Dynamic Link Prediction Performance Comparison with different diffusion messages.}
    \label{fig:messages}
\end{figure*}

Another interesting phenomenon is that the training time and convergence epochs of \textbf{LDG-AD-base} and \textbf{LDG} are close to each other. \textbf{DyRep-AD-base} takes 4x time per epoch than \textbf{DyRep}, but \textbf{DyRep-AD-base} only needs one epoch to converge, while \textbf{DyRep} takes 5 epochs, which means that the AD mechanism does not increase the training time in practice. The reason why the AD mechanism can reduce the number of epochs required for convergence is that one of the purpose of the repeated epochs in the training process is to propagate the delayed information through sample repetition. For example, in Fig.~\ref{fig:delay_issuse}, in the $i$-th training epoch $o^{t_2}$ will lead to the information updated at $t_4$, $t_5$ in the previous epoch to be propagated to node $b$ and thus to $e$ when $o^{t_6}$ occurs in the $i$-th training epoch.

\subsection{Impact of Diffusion Message}\label{sec:message}
In this section, we discuss the impact of diffusion message. We replace the diffusion message generation method in \textit{*-AD-base} with Eq.~\ref{eq:message_delta} (\textit{*-AD-delta}) and Eq.~\ref{eq:message_edge} (\textit{*-AD-edge}). Fig.~\ref{fig:messages} shows the dynamic link prediction performance comparison with different diffusion messages.

We find that \textit{*-AD-delta} have a significant performance decline on the Social Evolution dataset and a significant improvement on Github dataset (especially \textbf{LDG-AD-delta}), while \textbf{LDG-AD-edge} performs better on the Social Evolution dataset.

In the Social Evolution dataset, the average interval for a node to appear in an event is 35.43 much lower than 107.83 in the Github dataset. The frequent interaction of nodes makes the diffusion information $\mathbf{m}$ generated using Eq.~\ref{eq:message_delta} in Social Evolution close is very small, so it cannot diffuse enough information, resulting in poor performance in the Social Evolution dataset. While in Github dataset, the longer interval allows nodes to accumulate enough difference information for diffusion.

The generation of edge diffusion message (Eq.~\ref{eq:message_edge}) involves both interacting nodes in the event, therefore, during diffusion, if the number of common neighbors between two nodes is high, it will cause duplicate overlap of diffusion message and thus bring negative impact to the performance. In Social Evolution dataset, the average number of common neighbors of interacting nodes in an event is 2.08, while in the Github dataset is 6.02.

\subsection{Impact of Diffusion Hops}\label{sec:hop}
\begin{figure*}[!htbp]
    \centering
    \subfigure[MAR on Social Evolution]{
    \includegraphics[width=0.45\linewidth]{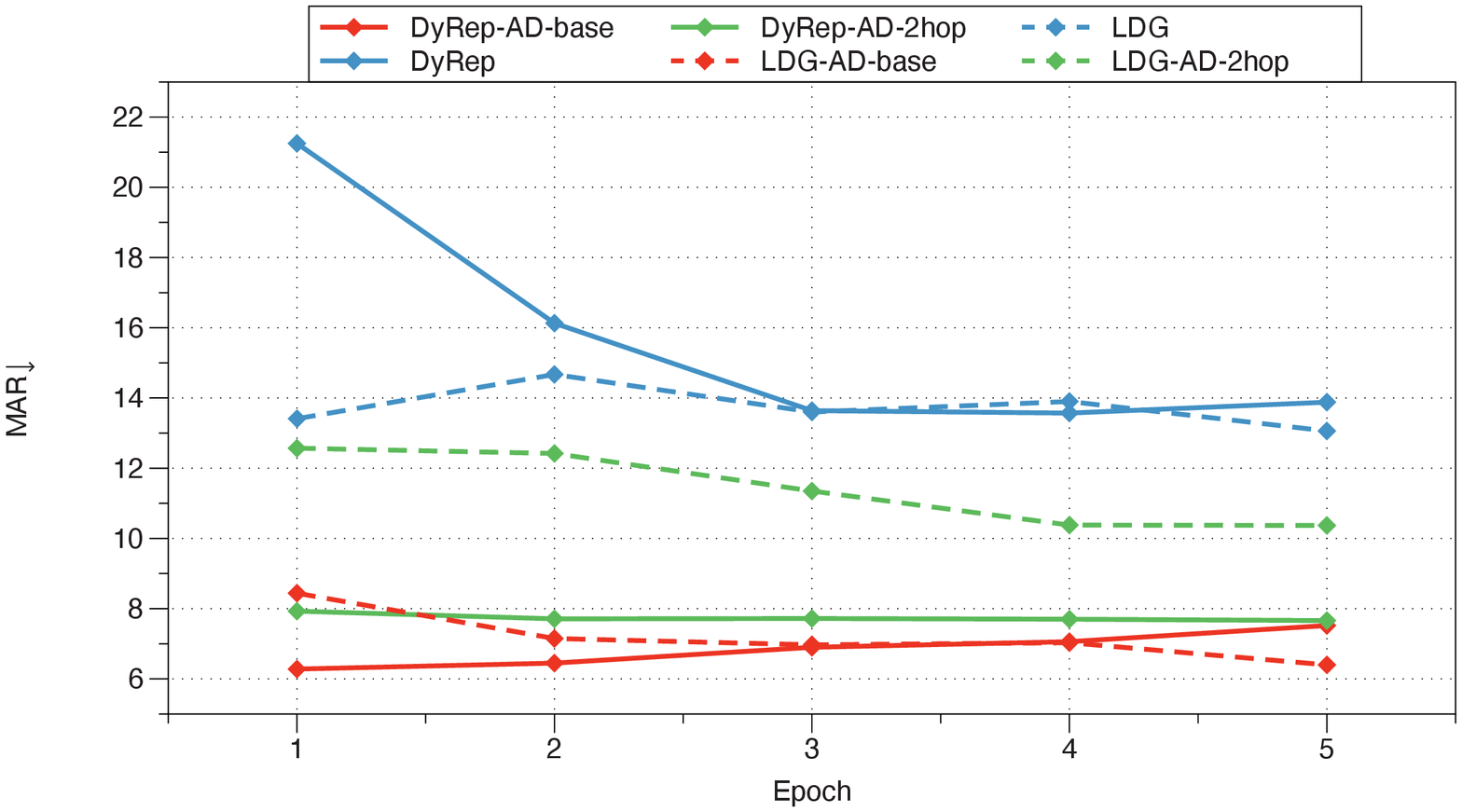}
    }
    \subfigure[MAR on Github]{
    \includegraphics[width=0.45\linewidth]{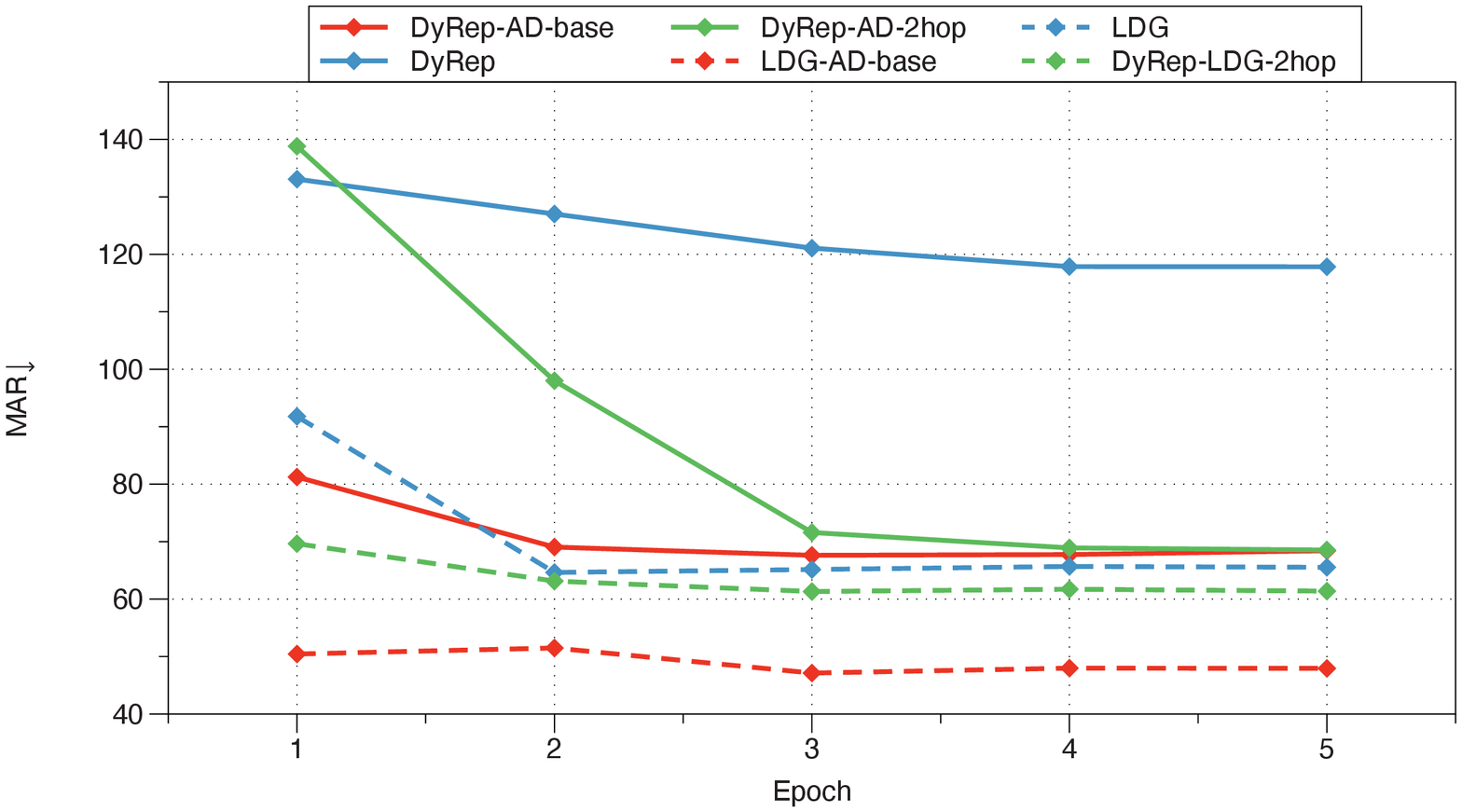}
    }
    \subfigure[HIT@10 on Social Evolution]{
    \includegraphics[width=0.45\linewidth]{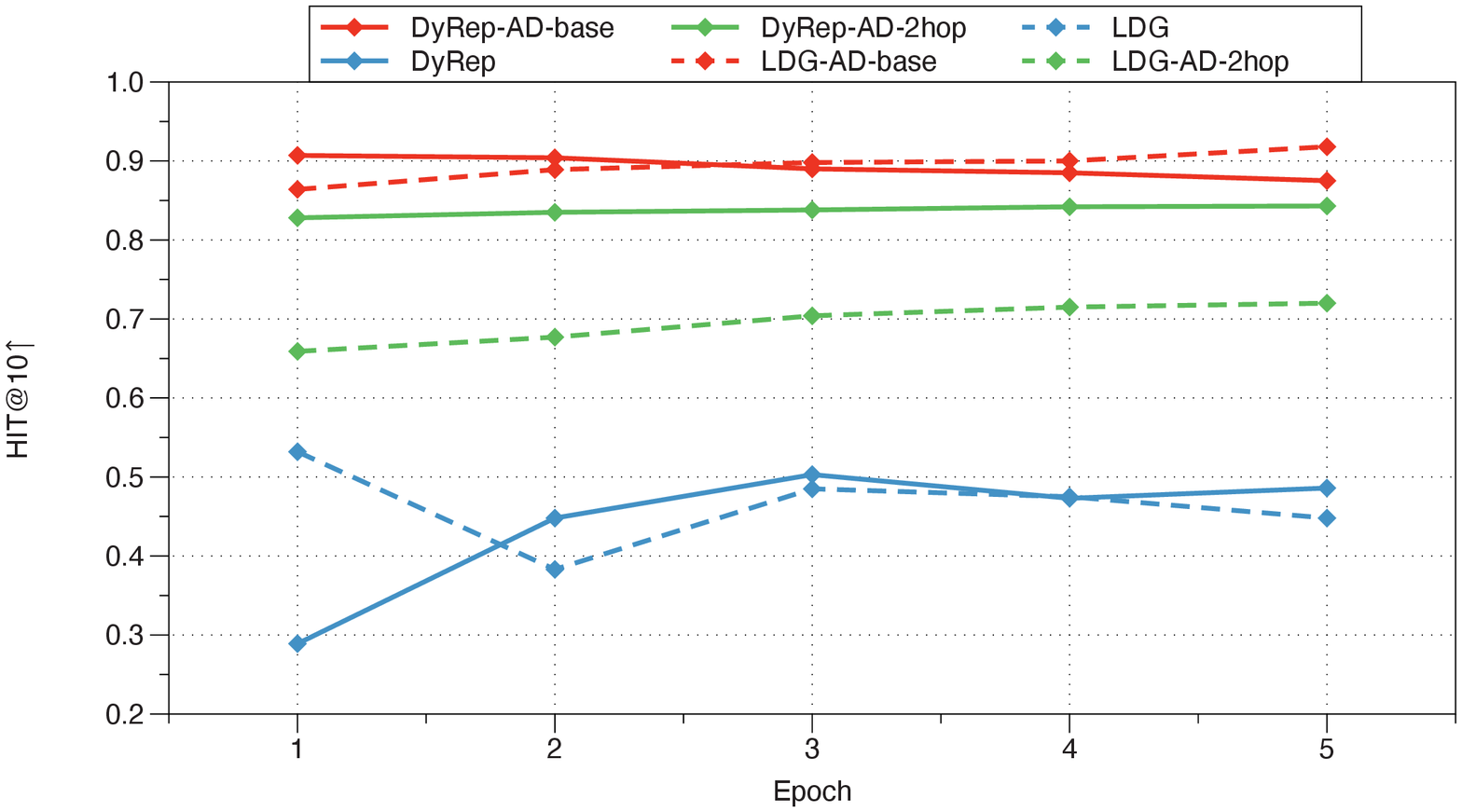}
    }
    \subfigure[HIT@10 on Github]{
    \includegraphics[width=0.45\linewidth]{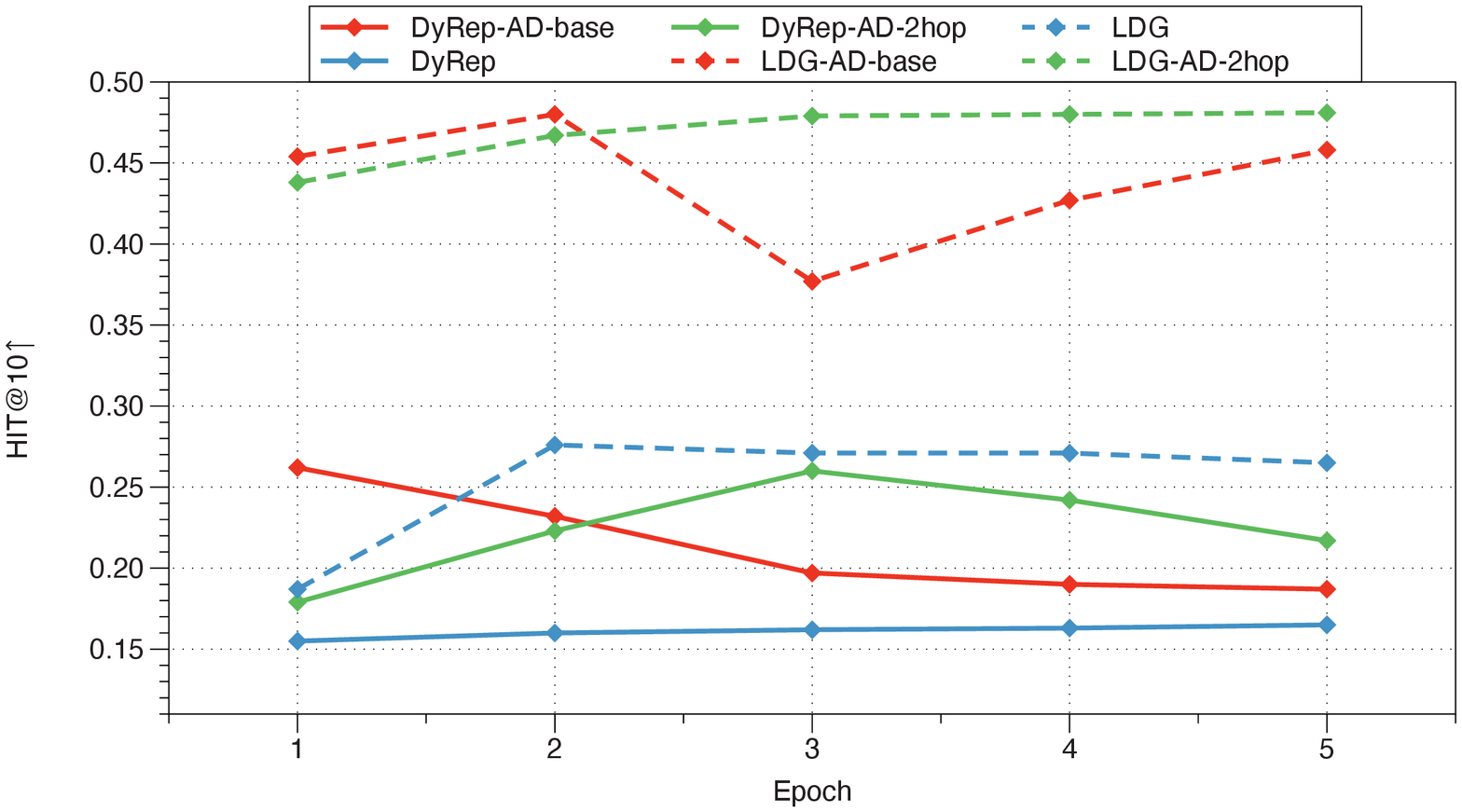}
    }
    \caption{Dynamic Link Prediction Performance Comparison with different diffusion hops.}
    \label{fig:hops}
\end{figure*}
Fig.~\ref{fig:hops} shows the dynamic link prediction performance comparison with different diffusion hops. Diffusing more hops on the Social Evolution and Github datasets doesn't result in improved performance, but instead has a negative effect. There are two main reasons for the negative effect:
\begin{itemize}
    \item The more hops of diffusion, the greater the risk and amount of noise introduced in the process of diffusion.
    \item The AD mechanism actually reduces the distance of information propagation between nodes. Compared to using only aggregation, the $i$-hop diffusion in the AD mechanism reduces the distance of information propagation by at least $i$. The average path length between nodes in the Social Evolution and Github datasets is 2.239 and 2.899, respectively. Therefore, 2-hop diffusion will make the information propagation distance less than 1 (0.239 and 0.899) which will result in the aggregated and diffused messages mixed together, thus producing a negative effect.
\end{itemize}
In addition, we find that as the number of diffusion hops increases, the training time consuming also increase dramatically, so we suggest that diffusion of 1-hop in practice can be a better balance between performance and training cost.

\subsection{Impact of Aggregation and Diffusion Nodes}\label{sec:who}
\begin{figure*}[!htbp]
    \centering
    \subfigure[MAR on Social Evolution]{
    \includegraphics[width=0.45\linewidth]{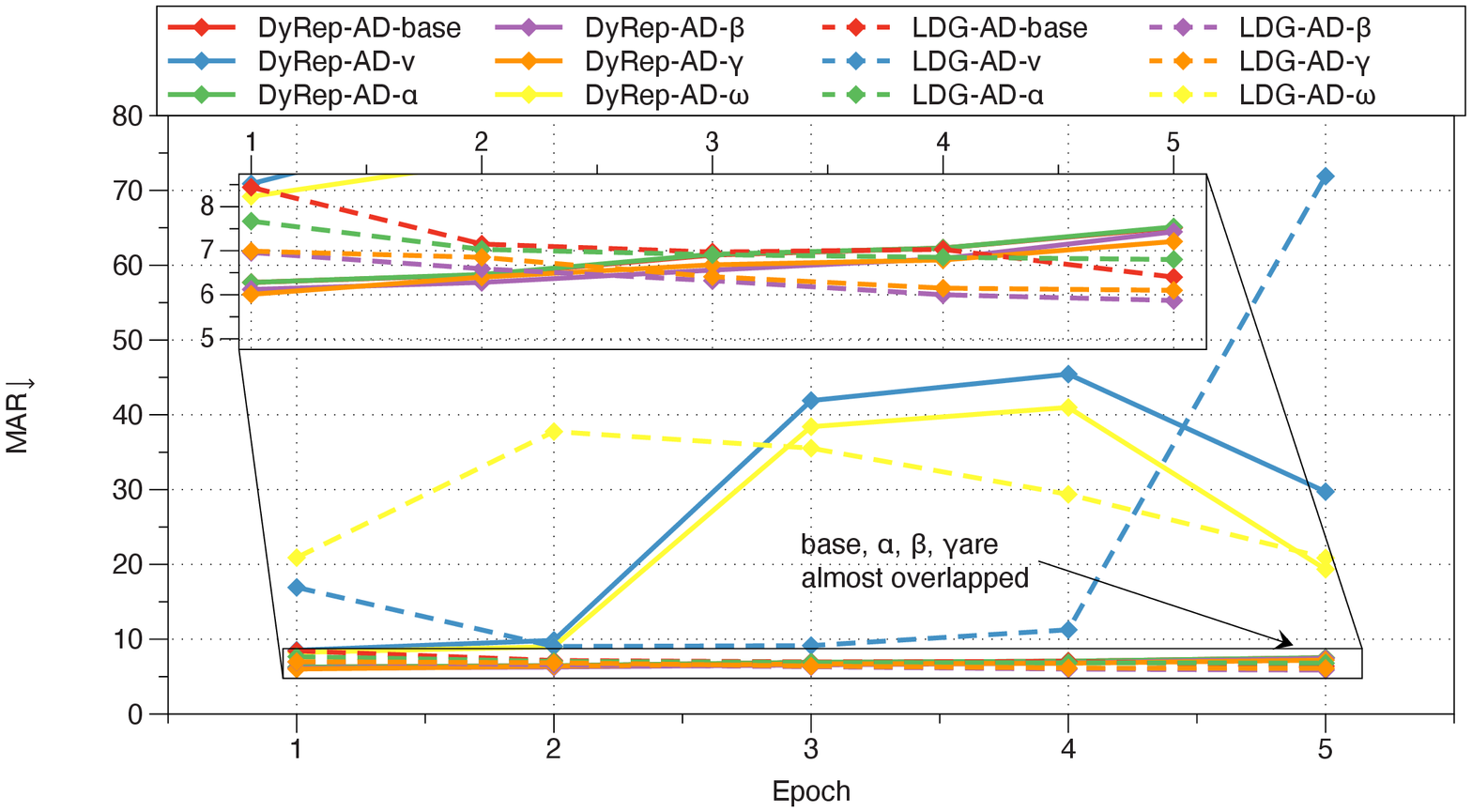}
    }
    \subfigure[MAR on Github]{
    \includegraphics[width=0.45\linewidth]{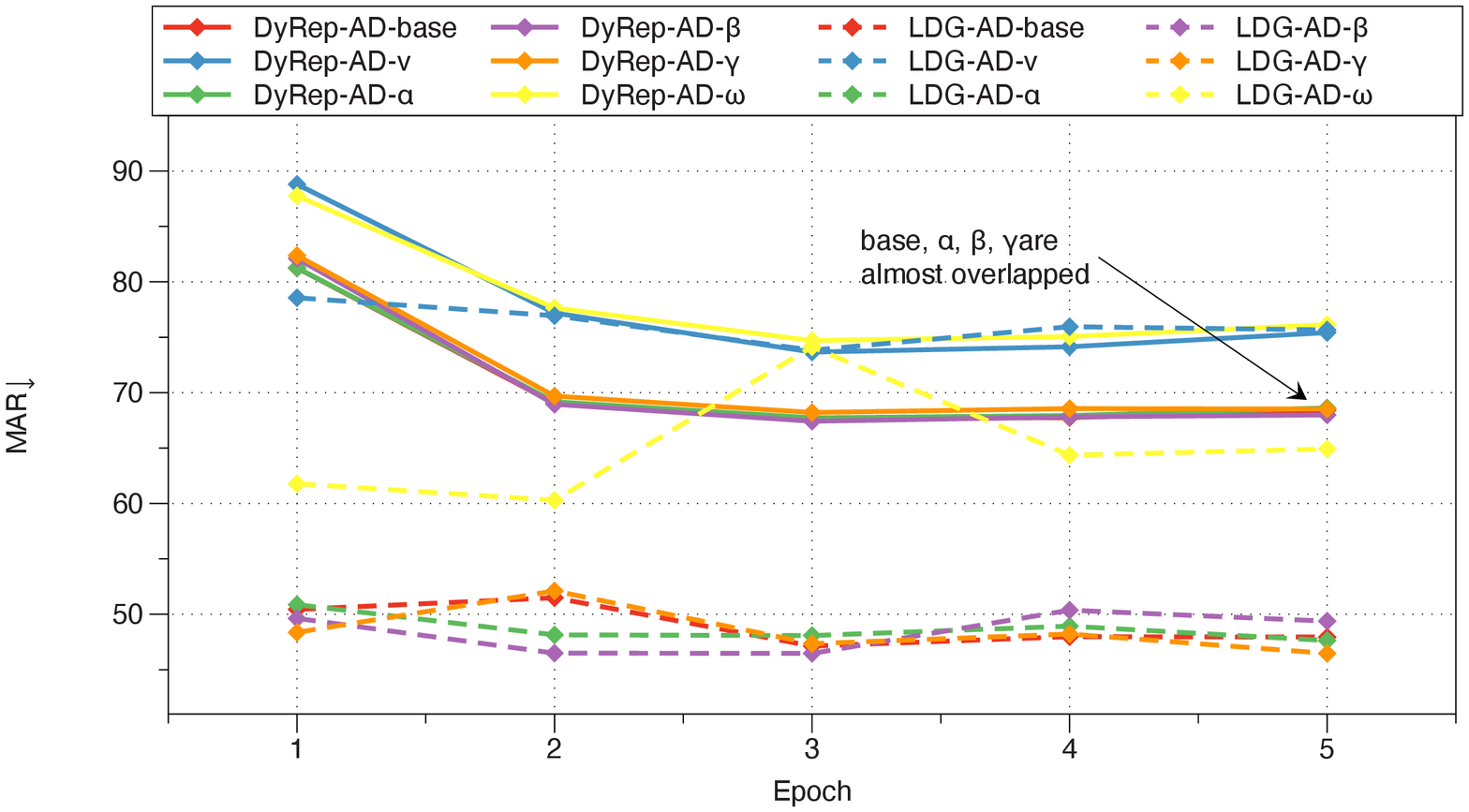}
    }
    \subfigure[HIT@10 on Social Evolution]{
    \includegraphics[width=0.45\linewidth]{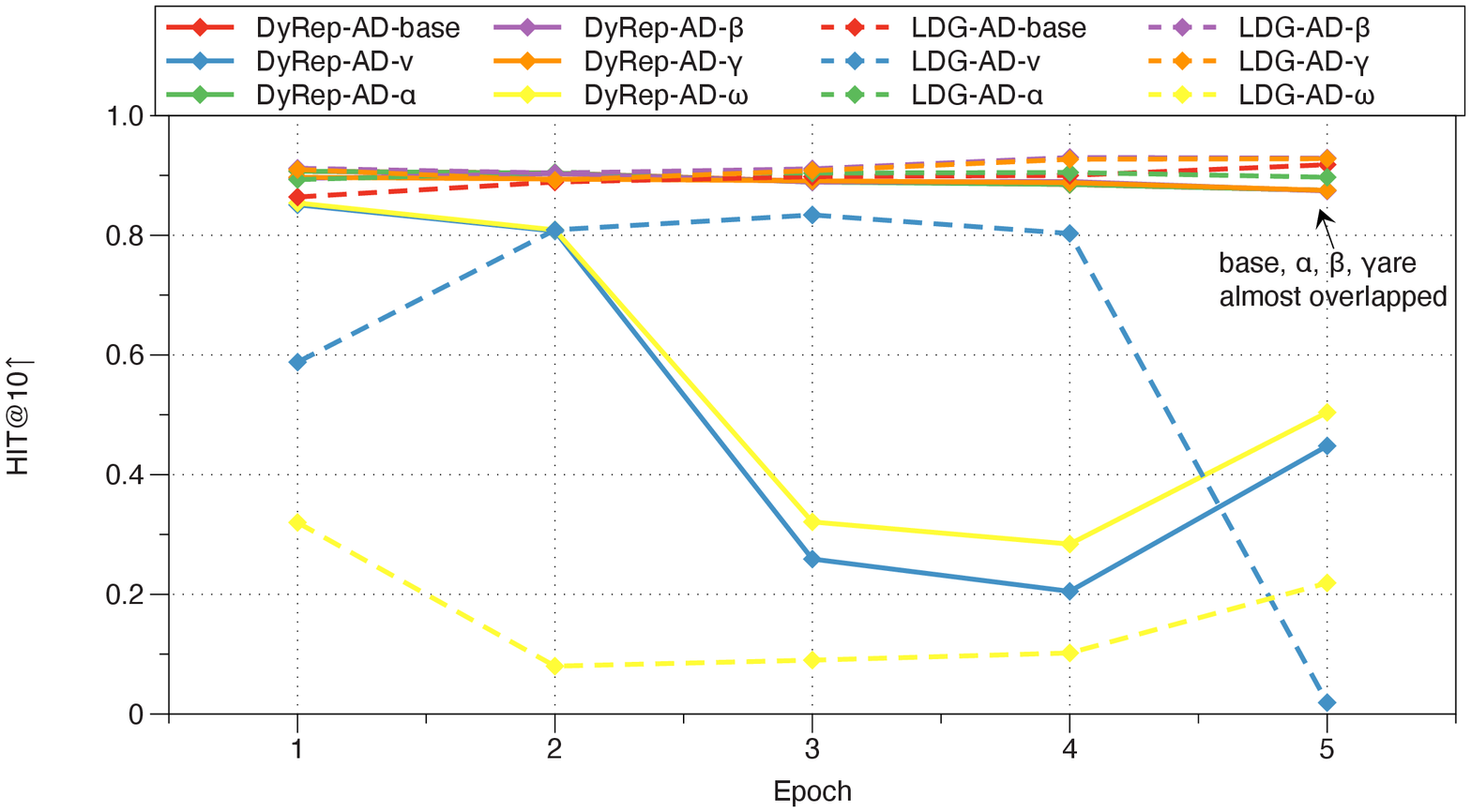}
    }
    \subfigure[HIT@10 on Github]{
    \includegraphics[width=0.45\linewidth]{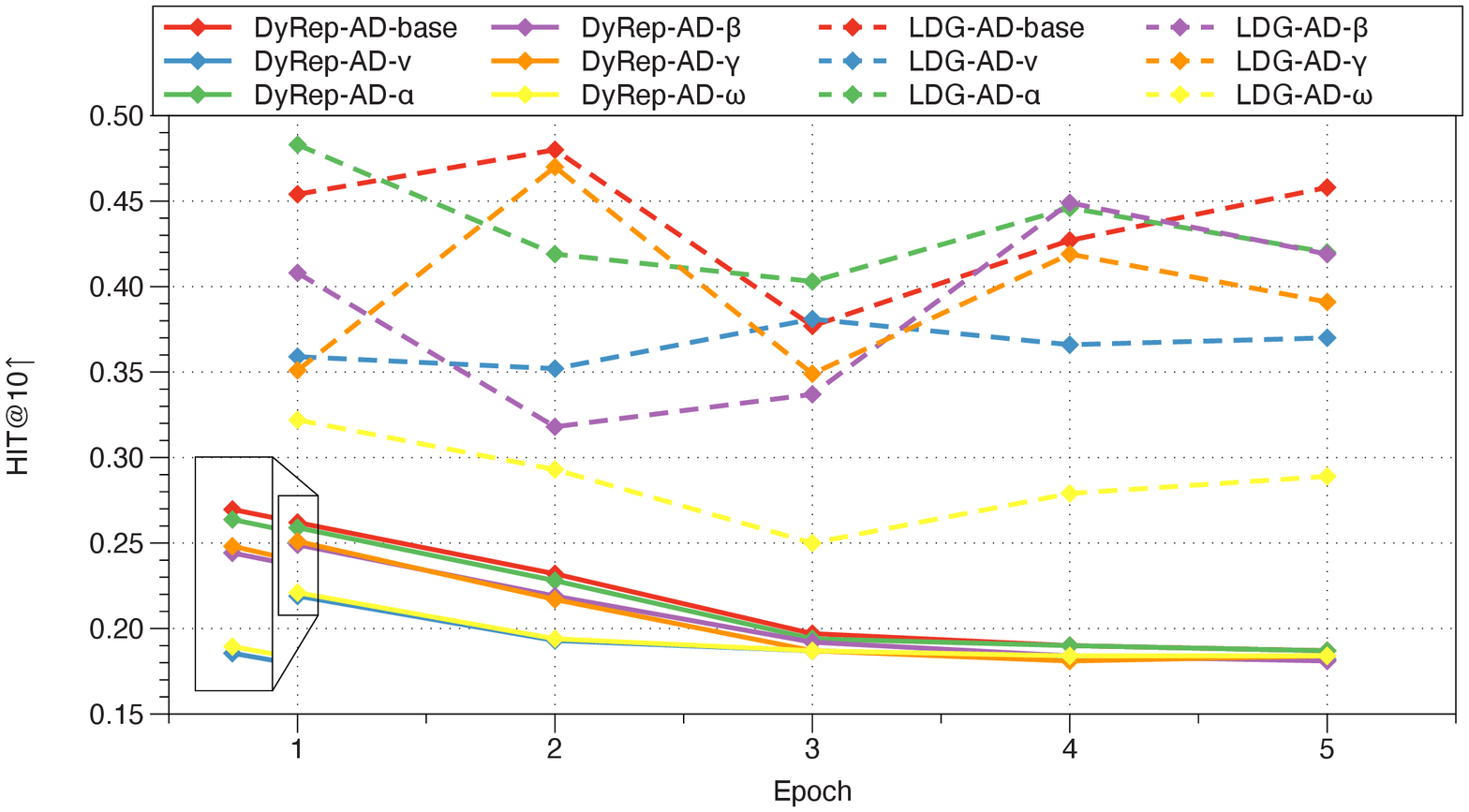}
    }
    \caption{Dynamic Link Prediction Performance Comparison with different diffusion nodes.}
    \label{fig:nodes}
\end{figure*}

In this section, we discuss the performance impact of different strategies for selecting aggregation and diffusion nodes.  We consider the following selection strategies and the results are shown in Fig.~\ref{fig:nodes}:
\begin{itemize}
    \item \textit{*-AD-v}: This strategy does not remove another interacting node $v$ when selecting diffusion nodes $\mathcal{N}_u^{d}$ of interacting node $u$: $\mathcal{N}_u^{d} = \{r: \mathbf{A}_{r,u}(\bar{t}) > 0 \}$. We find a significant performance drop in Fig.~\ref{fig:nodes}. This is due to $v$ has already obtained information about $u$ through aggregation step, and repeatedly obtaining information through diffusion will lead to negative effect.
    
    \item \textit{*-AD-$\alpha$}: This strategy randomly mask 20\% of the neighboring nodes of interacting node at the aggregation step but diffusion is still based on Eq.~\ref{eq:who_neighbor}. We find that \textbf{LDG-AD-$\alpha$} performs slightly better than \textbf{LDG-AD-base} on both Social Evolution and Github datasets, while \textbf{DyRep-AD-$alpha$} performs slightly worse than \textbf{DyRep-AD-base}.
    
    \item \textit{*-AD-$\beta$}: This strategy randomly mask 20\% of the neighboring nodes of interacting node at the diffusion step but still aggregate all neighbors. We find that \textbf{DyRep-AD-$\beta$} and \textbf{LDG-AD-$\beta$} slightly improved the MAR metrics on both Social Evolution and Github datasets, while \textbf{LDG-AD-$\beta$} significantly decreased the HIT@10 metric on Github.
    
    \item \textit{*-AD-$\gamma$}: This strategy randomly mask 20\% of the neighboring nodes of interacting node at both aggregation and diffusion step, and these two masks are independent. We observe no significant difference between \textit{*-AD-$\gamma$} and \textit{*-AD-base}. Compared with \textbf{LDG-AD-base}, \textbf{LDG-AD-$\gamma$} slightly improved the HIT@10 metrics on Social Evolution, while there is a significant gap on Github.

    \item \textit{*-AD-$\omega$}: This strategy mask 20\% of the earliest neighboring nodes of interacting node at both aggregation and diffusion step. We observe a significant drop in the performance of both MAR and HIT@10 on both Social Evolution and Github datasets.
\end{itemize}

\subsection{Impact of Diffusion Attention}\label{sec:attn}
\begin{table}[!hbtp]
\caption{Performance comparison of whether to use attention during diffusion.  \textcolor{blue}{\textbf{Blue bolded}} results denote best performance for DyRep-based model, \textcolor[rgb]{0.502,0,0.502}{\textbf{Purple bolded}} results denote best performance for LDG-based model.}\label{tab:attn}
\centering
\begin{tabular}{clcc} 
\hline
\multicolumn{1}{c}{DATASET}                                                        & MODEL         & MAR$\downarrow$                              & HIT\@10$\uparrow$                              \\ 
\hline
\multirow{4}{*}{\begin{tabular}[c]{@{}c@{}}SOCIAL\\EVOLUTION\end{tabular}}                                 & DyRep-AD-base & 6.28                                           & \textbf{\textcolor{blue}{0.907}}                \\
                                                                            & DyRep-AD-attn & \textbf{\textcolor{blue}{6.20}}                & 0.885                                           \\
                                                                            & LDG-AD-base   & \textbf{\textcolor[rgb]{0.502,0,0.502}{6.40}}  & \textbf{\textcolor[rgb]{0.502,0,0.502}{0.918}}  \\
                                                                            & LDG-AD-attn   & 6.63                                           & 0.907                                           \\ 
\hline \hline
\multirow{4}{*}{GITHUB} & DyRep-AD-base & \textbf{\textcolor{blue}{81.25}}               & \textbf{\textcolor{blue}{0.262}}                \\
                                                                            & DyRep-AD-attn & 84.78                                          & 0.261                                           \\
                                                                            & LDG-AD-base   & 51.49                                          & \textbf{\textcolor[rgb]{0.502,0,0.502}{0.480}}  \\
                                                                            & LDG-AD-attn   & \textbf{\textcolor[rgb]{0.502,0,0.502}{44.96}} & 0.427                                           \\
\hline

\end{tabular} 

\end{table}

In the work of GAT\cite{velivckovic2017graph}, it has been demonstrated that attention mechanisms have an important role in the aggregation process. And in this section, we explore whether existing attention has a positive effect on the diffusion process. Table~\ref{tab:attn} shows the performance comparison of whether to use attention during diffusion, where \textit{*-AD-attn} is the variation of \textit{*-AD-base} with replacing Eq.~\ref{eq:uniform} by Eq.~\ref{eq:attention}. We observed that \textit{*-AD-base} and \textit{*-AD-attn} each have strengths in different datasets and different evaluation metrics. Therefore, we believe that the attention mechanism is still meaningful in the diffusion process, and we will explore a more suitable attention mechanism for the diffusion process in our future work.

\subsection{Summary \& Suggestions}
Based on the above discussion, in this section, we summarize the impact factors in the AD mechanism and give some suggestions for using the AD mechanism in practice:
\begin{itemize}
    \item In generating the diffusion message, competitive performance can be obtained based on the most concise Eq.~\ref{eq:message_node}. Event (edge) message (Eq.~\ref{eq:message_edge}) is a good choice in case of few common neighbors between interacting nodes when focusing on HIT@10 metric.
    \item In practice, a remarkable performance can be obtained by diffusing 1-hop. When the average path length in the graph is too long, an appropriate increase in the number of hops can be considered.
    \item There is no need to filter the neighbors during aggregation, but there is a necessity to avoid propagate the diffusion message to another interacting node during the diffusion.
\end{itemize}

\section{Conclusion}
We introduce a novel aggregation-diffusion mechanism into the update of node embedding to extend the existing models with TPP-based DyRep and LDG as example. By using the AD mechanism, we get a huge improvement in all evaluation metrics on both Social Evolution and Github dataset compared to the original models. We also construct extensive experiments to explore the effects of different factors on the AD mechanism and give some suggestions for selecting suitable strategies of the aggregation and propagation process based on the graph properties.

In this paper, we have validated the effectiveness of the aggregation-diffusion mechanism mainly from experiments. In the further work, we will try to explain how the aggregation-diffusion mechanism works from the theoretical level.


%

\section*{Acknowledgment}
The research in this paper is partially supported by the National Key Research and Development Program of China (No 2018YFB1402500) and the National Science Foundation of China (61772155, 61832004, 61802089, 61832014).

\ifCLASSOPTIONcaptionsoff
  \newpage
\fi



\bibliographystyle{IEEEtran}
\bibliography{reference.bib}

\begin{thebibliography}{10}
\providecommand{\url}[1]{#1}
\csname url@samestyle\endcsname
\providecommand{\newblock}{\relax}
\providecommand{\bibinfo}[2]{#2}
\providecommand{\BIBentrySTDinterwordspacing}{\spaceskip=0pt\relax}
\providecommand{\BIBentryALTinterwordstretchfactor}{4}
\providecommand{\BIBentryALTinterwordspacing}{\spaceskip=\fontdimen2\font plus
\BIBentryALTinterwordstretchfactor\fontdimen3\font minus
  \fontdimen4\font\relax}
\providecommand{\BIBforeignlanguage}[2]{{%
\expandafter\ifx\csname l@#1\endcsname\relax
\typeout{** WARNING: IEEEtran.bst: No hyphenation pattern has been}%
\typeout{** loaded for the language `#1'. Using the pattern for}%
\typeout{** the default language instead.}%
\else
\language=\csname l@#1\endcsname
\fi
#2}}
\providecommand{\BIBdecl}{\relax}
\BIBdecl

\bibitem{scarselli2008graph}
F.~Scarselli, M.~Gori, A.~C. Tsoi, M.~Hagenbuchner, and G.~Monfardini, ``The
  graph neural network model,'' \emph{IEEE transactions on neural networks},
  vol.~20, no.~1, pp. 61--80, 2008.

\bibitem{kipf2016semi}
T.~N. Kipf and M.~Welling, ``Semi-supervised classification with graph
  convolutional networks,'' \emph{arXiv preprint arXiv:1609.02907}, 2016.

\bibitem{wang2016structural}
D.~Wang, P.~Cui, and W.~Zhu, ``Structural deep network embedding,'' in
  \emph{Proceedings of the 22nd ACM SIGKDD international conference on
  Knowledge discovery and data mining}, 2016, pp. 1225--1234.

\bibitem{skarding2020foundations}
J.~Skarding, B.~Gabrys, and K.~Musial, ``Foundations and modelling of dynamic
  networks using dynamic graph neural networks: A survey,'' \emph{arXiv
  preprint arXiv:2005.07496}, 2020.

\bibitem{trivedi2019dyrep}
R.~Trivedi, M.~Farajtabar, P.~Biswal, and H.~Zha, ``Dyrep: Learning
  representations over dynamic graphs,'' in \emph{International Conference on
  Learning Representations}, 2019.

\bibitem{knyazev2019learning}
B.~Knyazev, C.~Augusta, and G.~W. Taylor, ``Learning temporal attention in
  dynamic graphs with bilinear interactions,'' \emph{arXiv preprint
  arXiv:1909.10367}, 2019.

\bibitem{han2020graph}
Z.~Han, Y.~Ma, Y.~Wang, S.~Günnemann, and V.~Tresp, ``Graph hawkes neural
  network for forecasting on temporal knowledge graphs,'' in \emph{Automated
  Knowledge Base Construction}, 2020.

\bibitem{trivedi2017know}
R.~Trivedi, H.~Dai, Y.~Wang, and L.~Song, ``Know-evolve: Deep temporal
  reasoning for dynamic knowledge graphs,'' in \emph{International Conference
  on Machine Learning}.\hskip 1em plus 0.5em minus 0.4em\relax PMLR, 2017, pp.
  3462--3471.

\bibitem{madan2011sensing}
A.~Madan, M.~Cebrian, S.~Moturu, K.~Farrahi \emph{et~al.}, ``Sensing the"
  health state" of a community,'' \emph{IEEE Pervasive Computing}, vol.~11,
  no.~4, pp. 36--45, 2011.

\bibitem{seo2018structured}
Y.~Seo, M.~Defferrard, P.~Vandergheynst, and X.~Bresson, ``Structured sequence
  modeling with graph convolutional recurrent networks,'' in
  \emph{International Conference on Neural Information Processing}.\hskip 1em
  plus 0.5em minus 0.4em\relax Springer, 2018, pp. 362--373.

\bibitem{narayan2018learning}
A.~Narayan and P.~H. Roe, ``Learning graph dynamics using deep neural
  networks,'' \emph{IFAC-PapersOnLine}, vol.~51, no.~2, pp. 433--438, 2018.

\bibitem{niepert2016learning}
M.~Niepert, M.~Ahmed, and K.~Kutzkov, ``Learning convolutional neural networks
  for graphs,'' in \emph{International conference on machine learning}.\hskip
  1em plus 0.5em minus 0.4em\relax PMLR, 2016, pp. 2014--2023.

\bibitem{chen2019lstm}
J.~Chen, J.~Zhang, X.~Xu, C.~Fu, D.~Zhang, Q.~Zhang, and Q.~Xuan, ``E-lstm-d: A
  deep learning framework for dynamic network link prediction,'' \emph{IEEE
  Transactions on Systems, Man, and Cybernetics: Systems}, 2019.

\bibitem{zheng2020mathnet}
X.~Zheng, B.~Zhou, M.~Li, Y.~G. Wang, and J.~Gao, ``Mathnet: Haar-like wavelet
  multiresolution-analysis for graph representation and learning,'' \emph{arXiv
  preprint arXiv:2007.11202}, 2020.

\bibitem{sanchez2018graph}
A.~Sanchez-Gonzalez, N.~Heess, J.~T. Springenberg, J.~Merel, M.~Riedmiller,
  R.~Hadsell, and P.~Battaglia, ``Graph networks as learnable physics engines
  for inference and control,'' in \emph{International Conference on Machine
  Learning}.\hskip 1em plus 0.5em minus 0.4em\relax PMLR, 2018, pp. 4470--4479.

\bibitem{chang2016compositional}
M.~B. Chang, T.~Ullman, A.~Torralba, and J.~B. Tenenbaum, ``A compositional
  object-based approach to learning physical dynamics,'' \emph{arXiv preprint
  arXiv:1612.00341}, 2016.

\bibitem{manessi2020dynamic}
F.~Manessi, A.~Rozza, and M.~Manzo, ``Dynamic graph convolutional networks,''
  \emph{Pattern Recognition}, vol.~97, p. 107000, 2020.

\bibitem{jin2019recurrent}
W.~Jin, H.~Jiang, M.~Qu, T.~Chen, C.~Zhang, P.~Szekely, and X.~Ren, ``Recurrent
  event network: Global structure inference over temporal knowledge graph,''
  \emph{arXiv preprint arXiv:1904.05530}, 2019.

\bibitem{liben2007link}
D.~Liben-Nowell and J.~Kleinberg, ``The link-prediction problem for social
  networks,'' \emph{Journal of the American society for information science and
  technology}, vol.~58, no.~7, pp. 1019--1031, 2007.

\bibitem{hisano2018semi}
R.~Hisano, ``Semi-supervised graph embedding approach to dynamic link
  prediction,'' in \emph{International Workshop on Complex Networks}.\hskip 1em
  plus 0.5em minus 0.4em\relax Springer, 2018, pp. 109--121.

\bibitem{hamilton2017inductive}
W.~L. Hamilton, R.~Ying, and J.~Leskovec, ``Inductive representation learning
  on large graphs,'' \emph{arXiv preprint arXiv:1706.02216}, 2017.

\bibitem{tang2015line}
J.~Tang, M.~Qu, M.~Wang, M.~Zhang, J.~Yan, and Q.~Mei, ``Line: Large-scale
  information network embedding,'' in \emph{Proceedings of the 24th
  international conference on world wide web}, 2015, pp. 1067--1077.

\bibitem{grover2016node2vec}
A.~Grover and J.~Leskovec, ``node2vec: Scalable feature learning for
  networks,'' in \emph{Proceedings of the 22nd ACM SIGKDD international
  conference on Knowledge discovery and data mining}, 2016, pp. 855--864.

\bibitem{hochreiter1997long}
S.~Hochreiter and J.~Schmidhuber, ``Long short-term memory,'' \emph{Neural
  computation}, vol.~9, no.~8, pp. 1735--1780, 1997.

\bibitem{sherstinsky2020fundamentals}
A.~Sherstinsky, ``Fundamentals of recurrent neural network (rnn) and long
  short-term memory (lstm) network,'' \emph{Physica D: Nonlinear Phenomena},
  vol. 404, p. 132306, 2020.

\bibitem{kumar2019predicting}
S.~Kumar, X.~Zhang, and J.~Leskovec, ``Predicting dynamic embedding trajectory
  in temporal interaction networks,'' in \emph{Proceedings of the 25th ACM
  SIGKDD International Conference on Knowledge Discovery \& Data Mining}, 2019,
  pp. 1269--1278.

\bibitem{rossi2020temporal}
E.~Rossi, B.~Chamberlain, F.~Frasca, D.~Eynard, F.~Monti, and M.~Bronstein,
  ``Temporal graph networks for deep learning on dynamic graphs,'' \emph{arXiv
  preprint arXiv:2006.10637}, 2020.

\bibitem{ma2020streaming}
Y.~Ma, Z.~Guo, Z.~Ren, J.~Tang, and D.~Yin, ``Streaming graph neural
  networks,'' in \emph{Proceedings of the 43rd International ACM SIGIR
  Conference on Research and Development in Information Retrieval}, 2020, pp.
  719--728.

\bibitem{cox2020multivariate}
D.~R. Cox and P.~A.~W. Lewis, ``Multivariate point processes,'' in
  \emph{Contributions to Probability Theory}.\hskip 1em plus 0.5em minus
  0.4em\relax University of California Press, 2020, pp. 401--448.

\bibitem{kipf2018neural}
T.~Kipf, E.~Fetaya, K.-C. Wang, M.~Welling, and R.~Zemel, ``Neural relational
  inference for interacting systems,'' in \emph{International Conference on
  Machine Learning}.\hskip 1em plus 0.5em minus 0.4em\relax PMLR, 2018, pp.
  2688--2697.

\bibitem{mei2017neural}
H.~Mei and J.~Eisner, ``The neural hawkes process: a neurally self-modulating
  multivariate point process,'' in \emph{Proceedings of the 31st International
  Conference on Neural Information Processing Systems}, 2017, pp. 6757--6767.

\bibitem{farine2017dynamics}
D.~Farine, ``The dynamics of transmission and the dynamics of networks,''
  \emph{Journal of Animal Ecology}, vol.~86, no.~3, pp. 415--418, 2017.

\bibitem{artime2017dynamics}
O.~Artime, J.~J. Ramasco, and M.~San~Miguel, ``Dynamics on networks:
  competition of temporal and topological correlations,'' \emph{Scientific
  reports}, vol.~7, no.~1, pp. 1--10, 2017.

\bibitem{kingma2014adam}
D.~P. Kingma and J.~Ba, ``Adam: A method for stochastic optimization,''
  \emph{arXiv preprint arXiv:1412.6980}, 2014.

\bibitem{velivckovic2017graph}
P.~Veli{\v{c}}kovi{\'c}, G.~Cucurull, A.~Casanova, A.~Romero, P.~Lio, and
  Y.~Bengio, ``Graph attention networks,'' \emph{arXiv preprint
  arXiv:1710.10903}, 2017.

\end{thebibliography}
%

%







\end{document}